\documentclass[11pt]{article}

\usepackage[preprint]{acl}

\usepackage{times}
\usepackage{latexsym}

\usepackage[T1]{fontenc}

\usepackage[utf8]{inputenc}

\usepackage{microtype}

\usepackage{inconsolata}

\usepackage{graphicx}

\usepackage{amsmath} 
\usepackage{booktabs}
\usepackage{makecell}
\usepackage{tabularx}
\usepackage{float}

\title{Beyond a Single Sentiment Score: Topical and Affective Dimensions of Emoji Meaning on Twitter}

\author{
  \textbf{Rohan James}, 
  \textbf{Ioannis Christofilogiannis}, 
  \textbf{Yuuv Jauhari}, 
  \textbf{Shreya Shah}, \and
  \textbf{Gauri Pethkar} \\
  School of Computer Science and Statistics \\
  Trinity College Dublin, Ireland \\
  \texttt{\{jamesro, christoi, jauhariy, shshah, pethkarg\}@tcd.ie}
}

\begin{document}
\maketitle
\begin{abstract}
Emoji are widely treated as compact affective tokens whose meaning is summarised by a single sentiment score. We argue that this picture is incomplete. From a corpus of 16{,}575 English tweets and a per-emoji metric we call \emph{stability delta} (within-class minus between-class cosine similarity of sentence embeddings), we show that emojis used in tightly clustered topical contexts also exhibit \emph{more} polarised crowd-sourced sentiment, not less. Validated against the Kralj Novak et al.\ (2015) sentiment lexicon, the central correlation is Spearman $\rho = +0.361$ (Holm $p = 0.015$, 95\% bootstrap CI $[+0.10, +0.59]$); against the independent Ferr\'{e} et al.\ (2025) Spanish 13-emotion ratings, $\rho = +0.509$ ($p = 0.0003$). The result survives permutation testing, BCa bootstrap, leave-one-out jackknife, partial correlation controlling for emoji frequency, and Holm correction across a four-test family. Topical coherence and affective dispersion are distinct dimensions of emoji meaning. As a downstream check we report leakage-free five-fold cross-validation results for sarcasm classification on the Emoji Sarcasm Dataset; all three baselines (a gated multimodal architecture, fine-tuned BERTweet features, and fine-tuned RoBERTa features) score above 0.97 macro F1, indicating the dataset has saturating emoji-label cues. We frame the classifier as supporting evidence rather than the contribution.
\end{abstract}

\section*{Keywords}
emoji; emoji sentiment; sentiment analysis; affective computing; sarcasm detection; sentence embeddings; computational pragmatics; social media / Twitter; emoji semantics; sentiment lexicons; multimodal fusion

\section{Introduction}

The stance a person holds towards a topic comes through in the affect they convey. In speech, facial expressions, gestures, and prosody carry much of that affective load. Text strips those channels away. Emoji have evolved as a compensatory device: small graphical signals that re-introduce paralinguistic cues into written discourse. What started as decoration has become a meaningful linguistic tool.

Structural evidence that this evolution is systematic comes from the corpus itself. In the Emoji Sarcasm Dataset (ESD) ~\cite{subramanian2019esd}, 58.3\% of emojis appear at the end of tweets, 28.5\% in the middle, and only 13.2\% at the start, replicating the sentence-final dominance reported for WhatsApp~\cite{sampietro2020} and instant messaging~\cite{garrison2011} at social-media scale.

A monotonic sentiment gradient runs from VADER~\cite{hutto2014vader} compound score $+0.267$ at the start position to $-0.018$ at the end, and substitutive function rates, where the emoji replaces a terminal sentiment word rather than decorating existing text, rising from 76.5\% (start) to 91.6\% (end), consistent with the affective-punctuation account~\cite{provine1993, robus2020}. This structural regularity sharpens the central question: if emoji placement follows clear pragmatic conventions, why do topical coherence and affective consistency dissociate?

Most computational approaches treat each emoji as a token with a fixed affective scalar, often a single sentiment score in $[-1, +1]$. That representation collapses a richer structure. We argue, and provide empirical evidence, that an emoji's meaning unfolds along at least two independent axes: a \emph{topical} axis (does the emoji co-occur with consistent textual content) and an \emph{affective} axis (does the emoji evoke a consistent emotional response). The two axes are commonly assumed to align. They do not.

Our main contribution is a finding rather than a method. We define the \emph{stability delta} of an emoji as the difference between within-class and between-class mean cosine similarity of sentence embeddings of the surrounding text. Higher delta means an emoji appears in textually more coherent contexts. Across 62 emojis with at least 15 occurrences in the ESD, we find that stability delta correlates positively with two independent measures of affective dispersion:

\begin{itemize}
\item Crowd-sourced sentiment variance from the Kralj Novak et al.\ (2015) lexicon (n=59 matched, $\rho = +0.361$, Holm-corrected $p = 0.015$, 95\% bootstrap CI $[+0.10, +0.59]$).
\item Number of strongly evoked emotions from the Ferr\'{e} et al.\ (2025) Spanish 13-emotion database (n=47 matched, $\rho = +0.509$, $p = 0.0003$).
\end{itemize}

Both findings survive a battery of robustness checks: permutation tests, BCa bootstrap intervals, leave-one-out jackknife, partial correlation controlling for emoji frequency, and Holm correction over the family of tests. The two correlations come from different methodologies, different languages, and different annotator populations, yet point in the same direction.

We then ask whether multimodal fusion of text, emoji, and structural features improves a downstream sarcasm classifier on ESD. Under leakage-free five-fold cross-validation, our gated architecture (GEMA) reaches macro F1 0.985, BERTweet feature extraction with logistic regression reaches 0.974, and RoBERTa with logistic regression reaches 0.991. All three sit above 0.97, indicating that ESD has surface-level emoji-label correlations strong enough to saturate any reasonable classifier. We therefore frame the classifier as a supporting application rather than the contribution of this paper.

The paper is organised as follows. Section~\ref{sec:related} reviews related work. Section~\ref{sec:data} describes the corpus and external resources. Section~\ref{sec:method} formalises the stability delta and the validation battery. Section~\ref{sec:results} presents results, including external validation and the application to sarcasm classification. Section~\ref{sec:discussion} discusses implications. Section~\ref{sec:conclusion} concludes.

\section{Related Work}\label{sec:related}

\paragraph{Emoji as paralinguistic signals.}
~\cite{sampietro2020} showed that around 55\% of emoji in WhatsApp messages occupy sentence-final position, suggesting they often function as affective punctuation. ~\cite{robus2020} found via eye-tracking that sentence-final emoji slow first-pass reading times, consistent with emoji acting as integrative cues rather than decoration. ~\cite{garrison2011} reported similar placement patterns in instant messaging. Beyond placement, ~\cite{wijeratne2017emoji_similarity} demonstrated that emoji similarity derived from sense-disambiguated definitions correlates with human judgement.

At social-media scale, sentence-final dominance holds with a substantial effect: a chi-square test on the \textbf{ESD} Twitter partition yields $\chi^2 = 4039.03$, $p < 10^{-5}$, Cram\'er's $V = 0.494$ for the association between emoji position and sarcasm label. Prior placement studies focused on structural distributions~\cite{sampietro2020, garrison2011, robus2020}; we additionally show that sentiment varies monotonically with position and that substitutive function rates increase from start to end placements, establishing that \emph{where} an emoji sits predicts its pragmatic role as reliably as \emph{which} emoji it is.

\paragraph{Sentiment lexicons for emoji.}
~\cite{kraljevnovak2015} aggregated 1.6 million tweet-level annotations from 83 annotators into per-emoji probability distributions over (negative, neutral, positive) labels and a derived sentiment score. The lexicon underpins much of the downstream emoji sentiment literature. Recent work by ~\cite{ferre2025emojidis} extends this to a 13-emotion discrete annotation database for 112 emojis, rated by 763 Spanish speakers. ~\cite{czestochowska2022contextfree} crowd-sourced one-word descriptions for 1{,}289 emojis without textual context, providing an independent ambiguity benchmark.

\paragraph{Emoji and sarcasm.}
~\cite{subramanian2019esd} introduced the Emoji Sarcasm Dataset and a BiGRU baseline using GloVe and emoji2vec features, reporting up to 0.991 accuracy on the Twitter partition under a single 60/20/20 split. ~\cite{Sukhavasi2024} report comparable accuracy with parallel TF-IDF and emoji-vector pipelines fused into an LSTM-BERT classifier. ~\cite{helal2024contextual} and ~\cite{VITMAN2023121068} show that contextual encoders such as RoBERTa with auxiliary emotion or sarcasm pre-training reach state-of-the-art F1 on harder benchmarks like Mustard, iSarcasmEval, and SemEval-2018 Task 3.

\paragraph{Topical versus affective dimensions.}
Cross-lingual analyses by ~\cite{barbieri2018temporal_emoji} and others have noted that emoji usage shifts across time and platform, hinting that emoji meaning has structure beyond a single scalar. Work on context-free ambiguity~\cite{czestochowska2022contextfree} treats meaning variability as a property to measure rather than to summarise. To our knowledge, no prior work has explicitly tested whether topical and affective coherence dissociate at the emoji level. That is the gap this paper addresses.

\section{Data}\label{sec:data}

\paragraph{Primary corpus: ESD.}
We use the Emoji Sarcasm Dataset~\cite{subramanian2019esd} as the primary corpus. The Twitter partition contains 16,859 tweets, each containing at least one emoji and labelled binary for sarcasm. After English filtering, 16,587 tweets remain; after emoji normalisation and primary-emoji extraction, the final analytical corpus contains 16,575 tweets. Retaining emojis with at least 15 occurrences leaves 62 emojis covering 15,569 embedded tweets. The analytical corpus is 38.06\% sarcastic.

\paragraph{External resources for validation.}
Three external resources are matched against our 62 emojis:

\begin{itemize}
\item \textbf{Kralj Novak et al.\ (2015)}~\cite{kraljevnovak2015}, accessed via the \texttt{emosent-py} package. Per-emoji sentiment score and probability distribution over negative/neutral/positive labels. 59 of our 62 emojis match.
\item \textbf{Ferr\'{e} et al.\ (2025) Emoji-Dis}~\cite{ferre2025emojidis}, downloaded from OSF. Mean Likert ratings (1--5 scale) across 13 discrete emotion categories from 763 Spanish speakers, covering 112 emojis. 47 of our 62 emojis match.
\item \textbf{Czestochowska et al.\ (2022) Context-Free Ambiguity}~\cite{czestochowska2022contextfree}, downloaded from the EPFL D-Lab GitHub. Roughly 30 one-word descriptions per emoji from human annotators with no textual context, covering 1{,}289 emojis. 59 of our 62 emojis match.
\end{itemize}

We also use the EmojiNet sense inventory~\cite{wijeratne2017emoji_similarity} (62/62 matched) to provide a sanity check based on dictionary polysemy.

\section{Method}\label{sec:method}

\subsection{Pre-processing}
\label{sec:preprocessing}

Each tweet is lowercased and stripped of URLs, \texttt{@mentions}, and hashtag symbols (the underlying word is kept). Unicode emoji characters are replaced with whitespace to produce \texttt{clean\_text}, which has all emoji removed and serves as the input to all downstream NLP components, including VADER scoring and sentence embedding. A separate emoji extraction pass operates on the original raw text: skin-tone modifiers (U+1F3FB through U+1F3FF) and Variation Selector-16 (U+FE0F) are stripped so presentation variants map to the same canonical form. The English filter retains tweets with at least 85\% ASCII alphabetic characters, removing 272 non-English tweets from the 16{,}859 raw inputs.

\subsection{Emoji Placement Analysis}
\label{sec:method-placement}

Each tweet in the Twitter partition is classified by the position of its first emoji as \textit{start}, \textit{middle}, or \textit{end} based on its relative location within the tweet, using the \texttt{emoji} library~\cite{python_emoji} for emoji identification and NLTK~\cite{BirdLoperKlein2009} for sentence segmentation and part-of-speech tagging. Sentiment is scored on the emoji-free \texttt{clean\_text} field (Section~\ref{sec:preprocessing}) using VADER~\cite{hutto2014vader} as the primary scorer and TextBlob as a cross-check; neither scorer sees emoji characters.

An emoji is classified as \textit{substitutive} if it occupies a position where a sentiment-bearing lexical item would grammatically occur, specifically, sentence-finally with no following content word, or directly preceding terminal punctuation, and \textit{additive} otherwise. Word count preceding the first emoji serves as a reading-load proxy, operationalising the prediction that longer propositional context before the emoji raises integrative processing demands~\cite{robus2020}.

Two statistical tests assess structural associations. A chi-square test on the (position $\times$ sarcasm label) contingency table evaluates whether emoji placement is independent of sarcasm, with Cram\'er's $V$ as effect size. A Mann-Kendall trend test on VADER compound scores sorted by reading load tests for a monotonic sentiment-processing-load relationship.

\subsection{Stability Delta}

For each emoji $e$ with at least 15 occurrences, let $T_e$ be the set of tweets in which $e$ is the primary emoji. We sample up to 100 tweets from $T_e$ and 500 random tweets from $T \setminus T_e$. Sentence embeddings are computed on the emoji-free \texttt{clean\_text} field using \texttt{all-MiniLM-L6-v2}~\cite{reimers2019sbert}, which produces 384-dimensional L2-normalised vectors. We then compute:
\begin{align}
W_e &= \frac{1}{|T_e^{\text{samp}}|(|T_e^{\text{samp}}|-1)} \sum_{i \neq j} \cos(\mathbf{x}_i, \mathbf{x}_j) \\
B_e &= \frac{1}{|T_e^{\text{samp}}| \cdot 500} \sum_{i, k} \cos(\mathbf{x}_i, \mathbf{y}_k) \\
\Delta_e &= W_e - B_e
\end{align}
where $i, j$ index tweets containing $e$ and $k$ indexes the random comparison set. A high $\Delta_e$ means tweets sharing emoji $e$ are textually more similar to each other than to typical tweets carrying other emoji. We treat $\Delta_e$ as a per-emoji measure of \emph{topical} coherence.

\subsection{External Validation Targets}

For each external resource we compute a per-emoji scalar that measures \emph{affective} or \emph{semantic} dispersion, then correlate it with $\Delta_e$:

\begin{itemize}
\item \textbf{Kralj Novak sentiment variance}: variance of $(p_{\text{neg}}, p_{\text{neu}}, p_{\text{pos}})$.
\item \textbf{Kralj Novak sentiment score}: scalar in $[-1, +1]$.
\item \textbf{Number of strongly evoked emotions}: count of Ferr\'{e} emotion categories with mean rating $\geq 2.5$ (the original threshold; we report sensitivity at 2.0, 3.0, and 3.5).
\item \textbf{Emotion entropy}: Shannon entropy over the baseline-subtracted (ratings minus 1) distribution across the 13 Ferr\'{e} categories.
\item \textbf{Description entropy} and \textbf{number of unique descriptions}: per-emoji statistics over Czestochowska crowd-sourced labels.
\item \textbf{EmojiNet sense count} and \textbf{keyword count}: dictionary polysemy markers.
\end{itemize}

\subsection{Statistical Battery}

For the four primary correlations involving Kralj Novak data, we select $\Delta$ vs.\ Kralj Novak variance as the primary test, with the remaining three treated as a family of related tests subject to Holm correction. To assess robustness, we run a seven-test statistical battery comprising Spearman $\rho$ (with permutation $p$-values, percentile, and BCa 95\% bootstrap CIs), Kendall's $\tau_b$, leave-one-emoji-out jackknife ranges, and partial Spearman rank-residuals controlling for emoji frequency (full methodology, assumptions, and software details in Appendix~\ref{sec:appendix_stats}). External verification on Ferr\'{e} 2025, EmojiNet, and Czestochowska are reported separately as confirmatory analyses.

\subsection{Sarcasm Classifier (Application)}\label{sec:sarcasm_classifier}

We run three classifiers under identical stratified five-fold cross-validation (random state = 42) on the English-filtered Twitter partition ($n = 16{,}587$):

\begin{itemize}

    \item \textbf{GEMA}: A multi-branch model combining text (BiGRU over MiniLM, BERT \texttt{[CLS]}), emoji, and structural features via a sigmoid gate. Complete architectural details and hyperparameters are summarized in Table~\ref{tab:gema_specs}.
    
    \item \textbf{BERTweet (frozen)}: \texttt{[CLS]} embeddings from \texttt{vinai/bertweet-base} fed into logistic regression.
    
    \item \textbf{RoBERTa (frozen)}: \texttt{[CLS]} embeddings from \texttt{roberta-base} fed into logistic regression.

\end{itemize}

Two leakage paths in earlier versions of GEMA are explicitly fixed: VADER is now scored on emoji-free clean text (avoiding contamination from VADER's built-in emoji lexicon), and centroid distance features are recomputed within each training fold rather than once on the full dataset. We confirmed both fixes with unit tests.

\section{Results}
\label{sec:results}
\subsection{Emoji Placement}
\label{sec:results-placement}

Emojis in the ESD Twitter partition are distributed as 58.3\% end, 28.5\% middle, and 13.2\% start, establishing end-placement as the strong default and replicating sentence-final dominance observed in smaller corpora~\cite{sampietro2020, garrison2011}.

\begin{figure}[h]
  \centering
  \includegraphics[width=\linewidth]{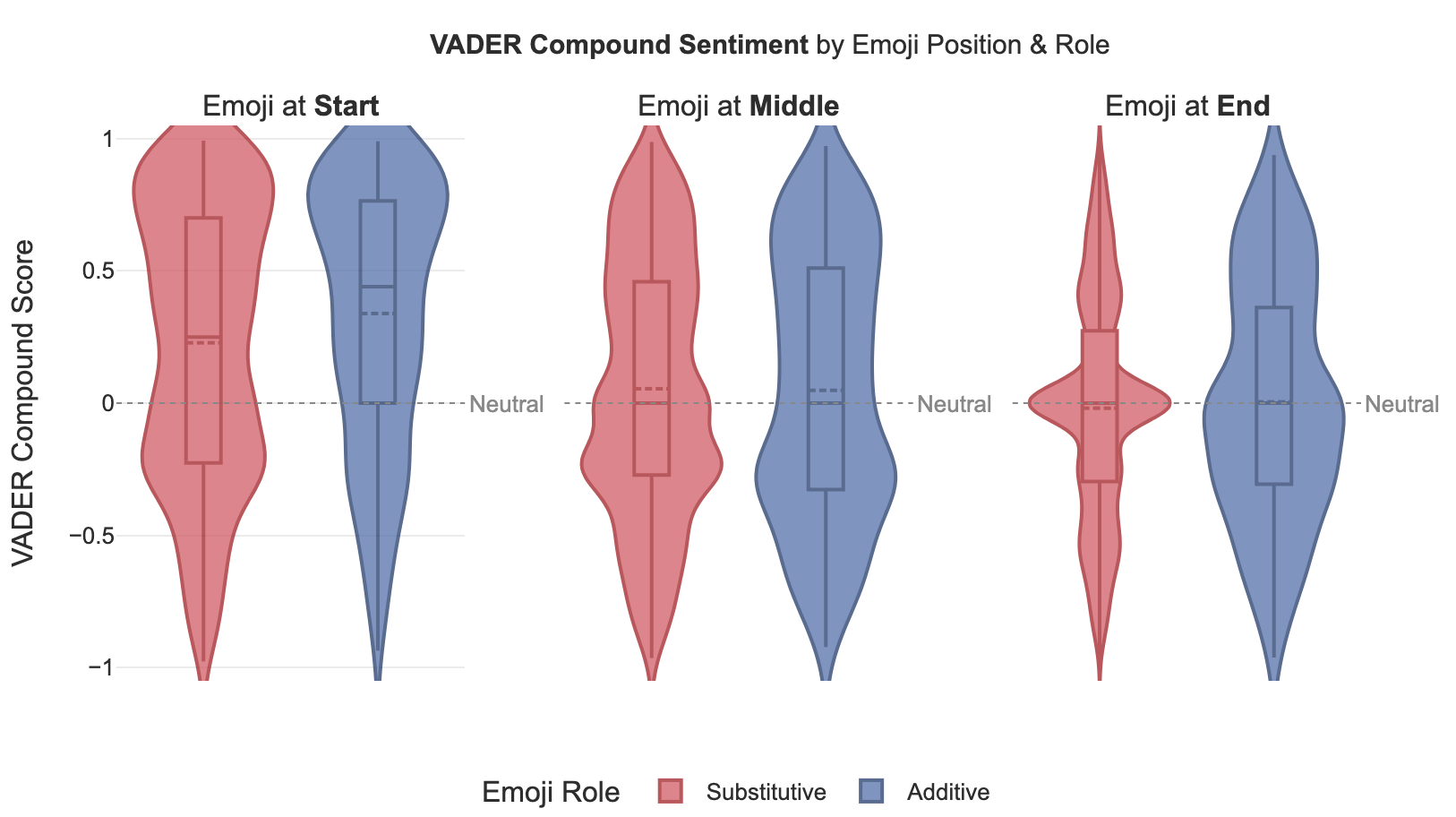}
  \caption{VADER compound score distributions by emoji position and role
  (substitutive vs.\ additive). Substitutive emojis produce more polarised
  distributions across all three positions.}
  \label{fig:violin-placement}
\end{figure}

Figure~\ref{fig:violin-placement} shows a clean monotonic sentiment gradient across both scorers. End-placed emojis, the largest group, score marginally negative on VADER, providing corpus-level evidence that trailing emojis are disproportionately associated with irony or sarcasm; start-placed emojis accompany genuinely positive text, signalling a tonal frame for the message before propositional content begins.

The reading-load proxy confirms this ordering: start-placed emojis are preceded by a mean of 1.78 words, middle-placed by 8.47, and end-placed by 10.95. End-positioned emojis thus follow a nearly complete proposition, meaning readers process the cognitive content before the affective signal, reflecting the eye-tracking evidence that sentence-final emoji slow integrative processing~\cite{robus2020}.

End-placed emojis are 91.6\% substitutive, middle-placed 81.7\%, and start-placed 76.5\%. Trailing emojis almost universally replace a concluding sentiment word; start-placed emojis show more additive usage. Complex punctuation rates run in the opposite direction: 31.9\% for start, 18.6\%
for middle, 4.2\% for end, suggesting that tweets beginning with an emoji exploit it as a framing device around which more syntactically complex, potentially ironic, structure is built.

Chi-square tests confirm both structural associations: position versus sarcasm label, $\chi^2 = 4039.03$, $p < 10^{-5}$, Cram\'er's $V = 0.494$; position versus VADER sentiment label, $\chi^2 = 1{,}267.79$, $p < 10^{-5}$, Cram\'er's $V = 0.196$. A Mann-Kendall trend test on VADER compound sorted by reading load yields $\tau = -0.0485$, $p < 10^{-5}$, indicating a weak but statistically reliable negative monotonic trend: longer preceding text is associated with slightly more negative sentiment.

\begin{figure}[h]
  \centering
  \includegraphics[width=\linewidth]{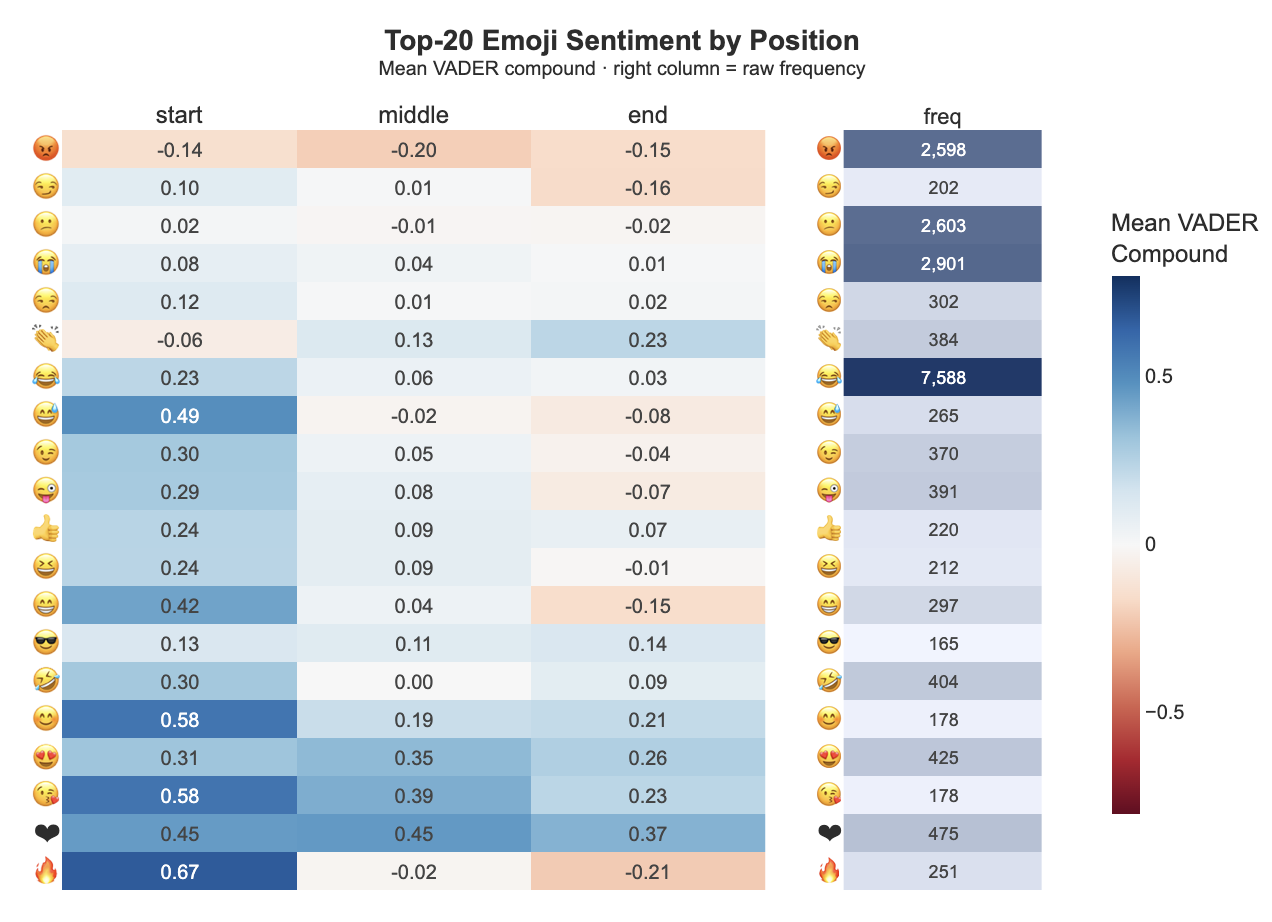}
  \caption{Mean VADER compound score for the top-20 emojis across positional
  contexts (start, middle, end). Right column shows raw frequency. Individual
  emojis exhibit marked affective shifts depending on position; where an emoji
  sits predicts its sentiment as strongly as which emoji it is.}
  \label{fig:heatmap-placement}
\end{figure}

Figure~\ref{fig:heatmap-placement} disaggregates these aggregate trends to the token level, plotting mean VADER compound scores for the 20 most frequent emojis across all three positional contexts alongside raw frequency counts. Two patterns are visible:

First, positional sentiment shifts are not uniform across emoji types.
Some tokens maintain a broadly consistent valence regardless of where they
appear in a tweet, while others undergo marked reversals between start and end positions. The magnitude and direction of the shift vary by token, which means that population-level placement norms: end-placed emojis, are marginally negative and these obscure substantial per-emoji heterogeneity.

Second, the compression effect is visible for high-frequency emojis.
Tokens appearing in thousands of tweets rather than tens show more muted
positional ranges, in-line with their broader, less topically constrained usage across many different conversational contexts. A rarely used emoji that appears almost exclusively in one register will track the sentiment of that register closely regardless of position; a high-frequency emoji used promiscuously across registers will average over them, flattening the positional gradient.

This token-level heterogeneity directly motivates the stability delta analysis that follows. Aggregate placement statistics capture the population-level gradient, but individual emoji behaviour is modulated by how topically coherent the contexts in which each symbol is deployed actually are. Where an emoji sits predicts its sentiment as reliably as which emoji it is at the population level; at the token level, that predictive power depends on the topical range of the emoji's typical usage, which is, the very quantity that stability delta is designed to measure.

\subsection{Stability Delta}

\begin{table}[h]
\centering
\small
\caption{Top and bottom emojis ranked by stability delta. ``Within'' and ``Between'' are mean pairwise cosine similarities within and across emoji classes.}
\label{tab:stability}
\begin{tabular}{lccc}
\toprule
\textbf{Emoji (description)} & \textbf{Within} & \textbf{Between} & \textbf{$\Delta$} \\
\midrule
Collision           & 0.409 & 0.172 & 0.237 \\
Fire                & 0.332 & 0.140 & 0.192 \\
Pointing-down       & 0.305 & 0.152 & 0.153 \\
\midrule
Astonished face     & 0.121 & 0.101 & 0.021 \\
Loudly crying face  & 0.143 & 0.124 & 0.019 \\
Confused face       & 0.124& 0.109& 0.014 \\
Face with tears of joy & 0.134& 0.123& 0.011 \\
Angry face          & 0.104& 0.107& $-$0.003 \\
\bottomrule
\end{tabular}
\end{table}

The 62 emojis range from $\Delta = +0.237$ for collision to $\Delta = -0.003$ for angry face, with mean $\Delta = 0.0766$ and 61 of 62 emojis positive (Table~\ref{tab:stability}). High-delta emojis appear in tightly clustered topical contexts; low-delta emojis are used across the full breadth of the corpus.

\subsection{Kralj Novak (2015) Validation}

The four primary correlations, with the full robustness battery, are reported in Table~\ref{tab:novak}.

\begin{table*}[h]
\centering
\small
\setlength{\tabcolsep}{3.8pt}
\caption{Kralj Novak (2015) validation, $n = 59$ matched emojis. Holm correction applied across the family of four tests. Significance markers refer to Holm-corrected $p$-values.}
\label{tab:novak}
\begin{tabular}{lccccccc}
\toprule
\textbf{Test} & \textbf{$\rho$} & \makecell{\textbf{Pct.\ CI}\\\textbf{(95\%)}} & \makecell{\textbf{BCa CI}\\\textbf{(95\%)}} & \textbf{Perm.\ $p$} & \textbf{$\tau_b$} & \makecell{\textbf{Partial $\rho$}\\\textbf{(freq)}} & \textbf{Holm $p$} \\
\midrule
$\Delta$ vs.\ Kralj Novak sentiment             & $+0.408$ & $[+0.16, +0.60]$ & $[+0.16, +0.60]$ & 0.0008 & $+0.258$ & $+0.423$ & 0.005 \\
$\Delta$ vs.\ Kralj Novak variance \textbf{(primary)} & $\mathbf{+0.361}$ & $[+0.10, +0.59]$ & $[+0.09, +0.58]$ & \textbf{0.005} & $+0.261$ & $+0.370$ & \textbf{0.015} \\
Sarcasm rate vs.\ $p(\text{neg})$         & $-0.286$ & $[-0.57, +0.02]$ & $[-0.55, +0.02]$ & 0.028  & $-0.227$ & $-0.280$ & 0.051 \\
Sarcasm rate vs.\ Kralj Novak sentiment         & $+0.291$ & $[-0.01, +0.57]$ & $[-0.01, +0.55]$ & 0.029  & $+0.234$ & $+0.289$ & 0.051 \\
\bottomrule
\end{tabular}
\end{table*}

The two stability-affect correlations survive Holm correction. For the primary test, $\Delta$ vs. Kralj Novak sentiment variance, the association is positive and robust: Spearman $\rho = +0.3610$, permutation $p = 0.0050$, percentile bootstrap CI $[+0.1029,+0.5872]$, BCa bootstrap CI $[+0.0885,+0.5813]$, Kendall $\tau_b = +0.2609$, partial $\rho$ controlling for emoji frequency $= +0.3700$, and Holm-corrected $p = 0.0149$. Leave-one-out jackknife yields $\rho$ values in $[+0.3327,+0.4119]$, so no single emoji drives the effect.

\begin{figure*}[h]
\centering
\includegraphics[width=0.46\textwidth]{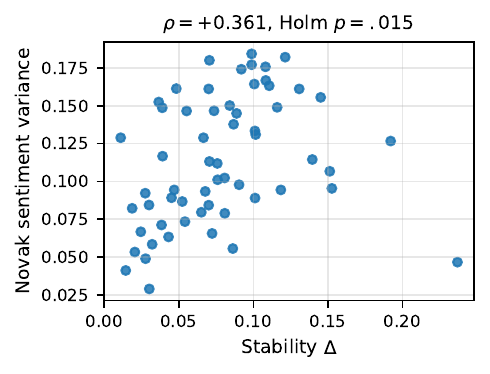}
\hfill
\includegraphics[width=0.46\textwidth]{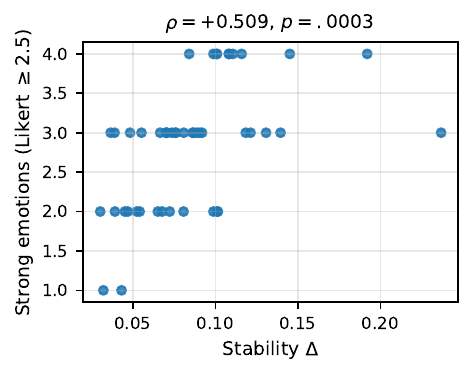}
\caption{Contextual stability is positively associated with affective dispersion. 
\textbf{Left:} stability $\Delta$ vs.\ Kralj Novak sentiment variance ($n=59$, Spearman $\rho=+0.361$, BCa$_{95}=[+0.089,+0.581]$, Holm $p=.015$). 
\textbf{Right:} stability $\Delta$ vs.\ number of strongly evoked emotions in Ferré et al. ($n=47$, $\rho=+0.509$, $p=.0003$). 
Emojis used in more topically coherent contexts evoke broader, not narrower, affective responses.}
\label{fig:stability-affect}
\end{figure*}

\begin{figure}[h]
\centering
\includegraphics[width=\linewidth]{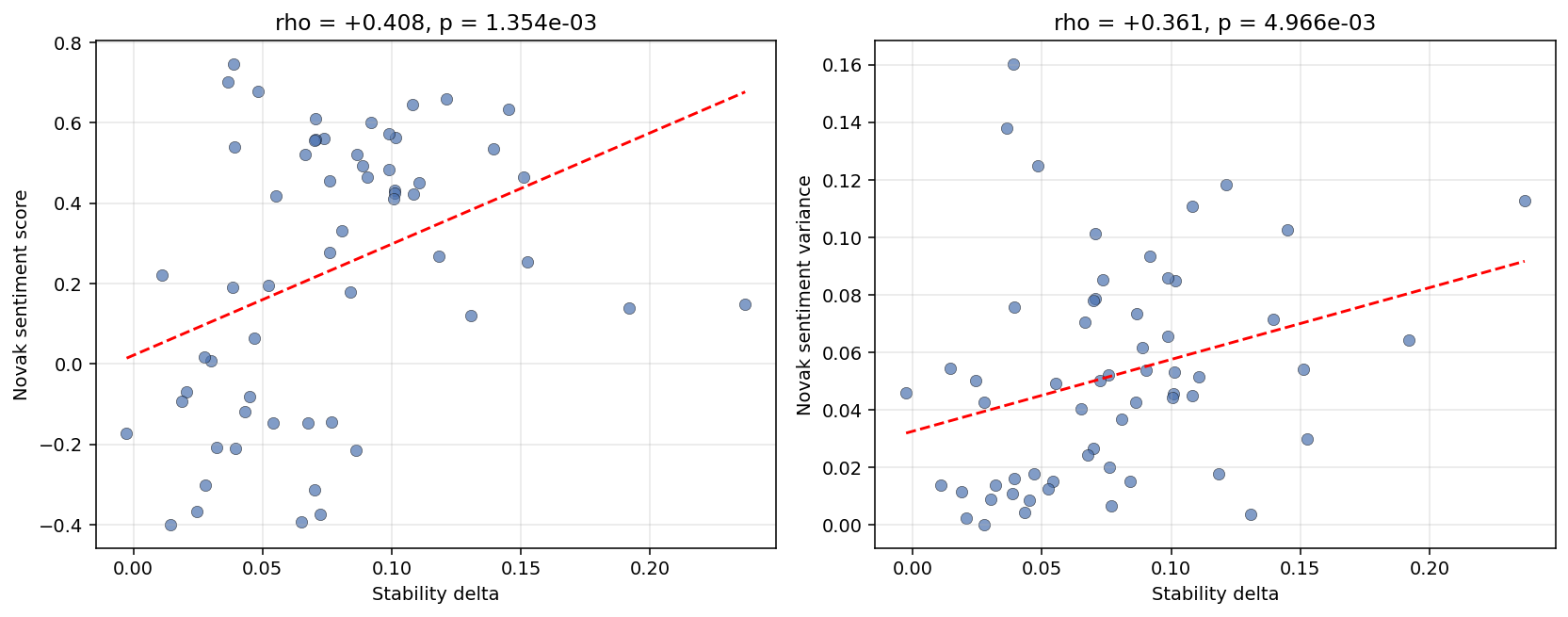}
\caption{Stability delta versus Kralj Novak sentiment score (left) and sentiment variance (right). Red lines are OLS fits. Both relationships are positive and Holm-significant.}
\label{fig:scatter}
\end{figure}

The two correlations involving sarcasm rate are positive in direction (sarcastic emojis lean toward positive Kralj Novak sentiment, away from negative probability mass) but do not survive Holm. We report them as exploratory. Figure~\ref{fig:scatter} shows the scatter plot for the primary tests. 


\subsection{External Corroboration: Ferr\'{e} 2025 Emoji-Dis}

A separate annotation effort by Ferr\'{e} et al.\ asked 763 Spanish speakers to rate 112 emojis on 1--5 Likert scales across 13 discrete emotion categories. We compute, per emoji, the number of categories with mean rating $\geq 2.5$ (``strongly evoked emotions''). 47 of our 62 emojis match.

\begin{table}[h]
\centering
\small
\caption{Stability delta vs.\ number of strongly evoked emotions in Ferr\'{e} 2025, with threshold sensitivity. $n = 47$ matched emojis throughout.}
\label{tab:ferre}
\begin{tabular}{lccc}
\toprule
\textbf{Likert threshold} & \textbf{$\rho$} & \textbf{$p$} & \textbf{Sig.} \\
\midrule
$\geq 2.0$ & $+0.606$ & $6.0 \times 10^{-6}$ & *** \\
$\geq 2.5$ \textbf{(primary)} & $\mathbf{+0.509}$ & $\mathbf{2.6 \times 10^{-4}}$ & *** \\
$\geq 3.0$ & $+0.458$ & $1.2 \times 10^{-3}$ & ** \\
$\geq 3.5$ & $+0.415$ & $3.7 \times 10^{-3}$ & ** \\
\bottomrule
\end{tabular}
\end{table}

The relationship is robust across all four threshold choices (Table~\ref{tab:ferre}). The strongest signal sits at lower thresholds, where emotion counts capture broader affective engagement; higher thresholds count only the most intensely evoked emotions.

\paragraph{Match-bias caveat.} The 47 Ferr\'{e}-matched emojis have a lower mean stability delta than the 15 unmatched emojis ($0.0674$ vs. $0.1054$; Mann--Whitney $p = 0.0376$). Because the unmatched emojis are not missing at random, we treat the Ferr\'{e} result as corroborating evidence rather than as a complete estimate over the full 62-emoji panel.

\subsection{Ambiguity and Polysemy: Sanity Checks}

The Czestochowska Context-Free Ambiguity dataset measures decontextualised emoji interpretation using description entropy and unique-word count ($n=59$), while EmojiNet measures dictionary polysemy via sense and keyword counts ($n=62$). Correlations between these metrics and $\Delta$ are weakly negative or near zero and not individually significant (full results in Appendix~\ref{sec:appendix_sanity_checks}), confirming that $\Delta$ captures contextualised topical coherence distinct from existing decontextualised ambiguity and dictionary polysemy measures.

\subsection{Polarity Alignment Between Text and Emoji}

We measure binary polarity alignment between the VADER compound on the emoji-free text and the Kralj Novak sentiment score for the primary emoji across tweets where Kralj Novak sentiment is available ($n = 15{,}398$). The alignment is 58.94\% overall, resting at 60.81\% for non-sarcastic tweets ($n = 10{,}131$) and 55.34\% for sarcastic tweets ($n = 5{,}267$). The non-sarcastic minus sarcastic alignment gap is only 5.47 percentage points, much smaller than a simple ``sarcasm equals polarity inversion'' account would predict.

\subsection{Sarcasm Classification (Application)}

\begin{table}[h]
\centering
\small
\caption{Five-fold stratified CV on the English Twitter partition ($n = 16{,}587$, mean $\pm$ SD). Full 5-metric evaluation reported in Appendix~\ref{sec:appendix_classifier}.}
\label{tab:classifier_summary}
\begin{tabularx}{\linewidth}{@{} l X X @{}}
\toprule
\textbf{Model} & \textbf{Macro F1} & \textbf{Sarc. F1} \\
\midrule
GEMA     & $.985{\pm}.007$ & $.982{\pm}.008$ \\
BERTweet & $.974{\pm}.002$ & $.967{\pm}.002$ \\
RoBERTa  & $.991{\pm}.001$ & $.988{\pm}.001$ \\
\bottomrule
\end{tabularx}
\end{table}

Table~\ref{tab:classifier_summary} reports classifier performance under identical splits (complete details in Appendix~\ref{sec:appendix_classifier}). All models exceed a macro F1 of 0.97. RoBERTa with a frozen \texttt{[CLS]} representation and a simple linear head slightly outperforms our gated multimodal architecture. We interpret this not as a triumph of frozen RoBERTa, but as evidence that ESD's emoji-label correlations make the classification task near-trivial, causing reasonable encoders to reach the performance ceiling. We retain the classifier solely as supporting evidence that emoji and structural features carry signal, rather than making architectural contribution claims.

\section{Discussion}
\label{sec:discussion}

The headline result is that stability delta correlates positively, not negatively, with affective dispersion. An emoji used in textually concentrated contexts also evokes a wider rather than narrower spread of crowd-sourced affect. We emphasise the agreement between two methodologically dissimilar resources: Kralj Novak's English tweet-level sentiment annotations~\cite{kraljevnovak2015} and Ferr\'{e}'s Spanish discrete-emotion Likert ratings~\cite{ferre2025emojidis}. Different annotators, different languages, different rating instruments, same direction.

Two interpretations are compatible with the data and we cannot fully separate them here. The first is that high-stability emojis tend to occupy specific cultural niches with strong affective valence (the fire emoji in praise contexts, the collision emoji in dramatic announcements) and the same niche supports both positive and negative affective use. The second is that crowd-sourced affective annotations are themselves polarised when an emoji has a clear cultural role, because annotators read the role rather than rate a neutral averaged perception. Either way, the practical implication is the same: representing emoji affect with a single sentiment scalar collapses the structure that the data actually carry.

\paragraph{Placement is systematic; semantics and affect still dissociate.}
The placement results establish that emoji use is far from arbitrary at the structural level. End-placement dominates at 58.3\% with Cram\'er's $V = 0.494$ linking position to sarcasm label - a medium-to-large effect by conventional standards~\cite{cohen1988statistical}, and substitutive function reaches 91.6\% for end-placed emojis, confirming that trailing emojis replace, rather than decorate, a concluding sentiment word. This structural regularity makes the semantic finding more striking, not less: even within the set of end-placed, substitutive emojis, the most pragmatically constrained subgroup - topical coherence (stability delta) and affective consistency (Kralj Novak variance) dissociate. The Cram\'er's $V = 0.196$ for position versus VADER label, and the 5.5 percentage-point polarity alignment gap between sarcastic and non-sarcastic tweets, jointly suggest that emoji-text sentiment mismatch is a real but weak global signal; position mediates part of that mismatch, but neither position nor mismatch alone accounts for sarcasm in this corpus. Single-scalar sentiment representations conflate these dimensions; the stability delta and the placement gradient together point toward multi-axis emoji representations as a productive direction for future work.

The negative results matter too. The Czestochowska context-free entropy and EmojiNet sense count correlations are weak and non-significant. This is consistent with our framing: $\Delta$ is a measure of \emph{contextual} usage stability, not of the lexicographer's enumeration of senses or of crowd interpretation in isolation. We treat the EmojiNet and Czestochowska results as evidence of construct distinctness rather than as failures of our metric. 

\paragraph{Polarity alignment is a weak sarcasm signal.} The 5.5-point gap in alignment between sarcastic and non-sarcastic tweets is small. This complicates the long-standing intuition that emoji-text polarity inversion is the defining feature of irony. Sarcasm in ESD is signalled by surface cues (specific emoji-phrase combinations) much more than by clean polarity inversion.

\paragraph{The classifier saturates.} All three baselines exceed 0.97 macro F1. We checked our work for the obvious explanations: leakage in the centroid feature, contamination of VADER scoring by the emoji lexicon, vocabulary built on the test fold. All three are demonstrably absent. The remaining explanation is that ESD's strong emoji-label correlations make the task near-trivial under any reasonable encoder. We therefore stop short of claiming architectural advantages for our gated fusion. A meaningful classifier comparison would require a harder benchmark such as iSarcasmEval~\cite{abu-farha2022isarcasm} or SemEval-2018 Task 3~\cite{vanhee2018semeval} where state-of-the-art F1 sits in the 0.6 range.

\section{Conclusion}
\label{sec:conclusion}

We presented the stability delta, a simple per-emoji metric of textual context coherence derived from sentence embeddings, and showed that it correlates \emph{positively} with two independent measures of affective dispersion: Kralj Novak (2015) sentiment variance ($\rho = +0.361$, Holm $p = 0.015$) and Ferr\'{e} (2025) number of strongly evoked emotions ($\rho = +0.509$, $p = 0.0003$). The result survives a battery of robustness checks. Topical coherence and affective valence are distinct dimensions of emoji meaning; representing emoji affect with a single sentiment scalar discards real structure. We provide a leakage-free five-fold evaluation of multimodal sarcasm classification on ESD as supporting context, and observe that the dataset is too easy to differentiate architectures meaningfully. Code, data-processing scripts, and validation outputs are provided as anonymised supplementary material.

Three directions are most promising. First, validation of stability delta on harder sarcasm benchmarks (iSarcasm, SemEval-2018 Task 3) where the affective-vs-topical distinction may yield architectural gains. Second, extension of the analysis to multi-platform timestamped corpora to test temporal generalisation properly. Third, exploring whether emoji embeddings learned with explicit affective-topical disentanglement (two heads, one for context, one for affect) outperform single-vector representations on downstream affective tasks.

\section*{Limitations}

The 62-emoji panel is small and the Kralj Novak sample at $n = 59$ is at the lower end of what supports rank correlation analysis, despite our best efforts at robust inference. The Ferr\'{e} match is biased toward lower-stability emojis (Section~\ref{sec:results}), so the $\rho = +0.509$ is a conservative estimate. We work in English on the topical side and rely on Spanish-rated emotions for one of the two external resources; cross-cultural transfer is an open question. We do not include a temporal analysis: a separate exploratory comparison between ESD Twitter and a 2025 BlueSky scrape was confounded by a 65-fold size imbalance, platform effects, and register differences (BlueSky data was seeded from \texttt{\#sarcasm} hashtags). We mention temporal drift only as future work. 

We release no new annotated corpus. The artefacts we share are derived per-emoji aggregates in CSV and JSON, so a formal datasheet~\cite{gebru2021datasheets} does not apply. These aggregates are structured tabular data tied to public DOI-tracked sources, in line with FAIR principles~\cite{Wilkinson2016The} for derived statistics.

\section*{Ethical Considerations}
Treating emoji affect as a single fixed score may trigger negative impacts when that score feeds into a downstream pipeline. A moderation system built this way will misread ironic intent, and the misreads will fall harder on figurative styles common in non-native and dialect varieties. A false sarcasm flag misroutes a moderation queue; a missed one lets ironic harm through. Representing topical and affective dimensions as separate axes, the change this paper recommends, narrows that failure mode.

All data we analyse is public and anonymised. Usernames are stripped from any redistributed slices, and no individual users are identified. Consent is inherited from the public-API terms of the original Twitter scrape and from the open-access licences of Kralj Novak, Ferr\'{e}, EmojiNet, and Czestochowska. We redistribute only derived per-emoji aggregates, never raw tweets. The Emoji Sarcasm Dataset contains offensive language typical of sarcastic discourse online; we do not curate or republish it.

Claude was utilized to assist with language refinement, editorial polishing of the manuscript and code generation for testing. All ideas, hypotheses, claims, analyses, code, results, citations, and final text were verified by the authors.

\bibliography{custom}

\appendix

\section{Statistical Battery Methodology and Assumptions}
\label{sec:appendix_stats}

To evaluate the stability and significance of the primary correlations involving Kralj Novak data, we subject each relationship to a seven-test statistical battery (10,000 resamples/permutations where applicable):

\begin{enumerate}
    \item \textbf{Spearman Rank Correlation ($\rho$):} Evaluates monotonic relationships using both asymptotic and empirical permutation $p$-values (10,000 permutations).
    \item \textbf{Percentile Bootstrap 95\% CI:} Non-parametric confidence intervals constructed over 10,000 resamples.
    \item \textbf{BCa Bootstrap 95\% CI:} Bias-corrected and accelerated confidence intervals (10,000 resamples) to adjust for skewness and acceleration in the sample distribution.
    \item \textbf{Kendall's $\tau_b$:} Serves as a rank-based sensitivity check on distributional shape and tied ranks.
    \item \textbf{Leave-One-Emoji-Out Jackknife:} Assesses high-leverage outliers by recording the range $[\rho_{\min}, \rho_{\max}]$ when each emoji is systematically omitted once.
    \item \textbf{Partial Spearman Correlation:} Controls for confounding emoji frequency $|T_e|$ using rank-residual analysis.
    \item \textbf{Holm Correction:} Adjusts $p$-values for family-wise error rate across the four-test family.
\end{enumerate}

\paragraph{Statistical Assumptions and Validation}
The battery assumes monotonic rank relationships between variables, exchangeability of observations under the null hypothesis for permutation testing, and independence across individual emoji entries. The partial Spearman procedure additionally assumes that rank residuals on emoji frequency exhibit approximate linearity; visual inspection confirmed that minor departures from linearity were negligible relative to small-$n$ sample variances, which are explicitly addressed by the BCa bootstrap intervals.

\section{Decontextualised Ambiguity and Polysemy Sanity Checks}
\label{sec:appendix_sanity_checks}

Table~\ref{tab:sanity_full} reports the sanity checks evaluating correlations ($\rho$) between our contextualised coherence metric ($\Delta$) and existing measures of decontextualised ambiguity \cite{czestochowska2022contextfree} and dictionary polysemy \cite{wijeratne2017emoji_similarity}. 

\begin{table*}[t]
\centering
\small
\caption{Full five-fold stratified cross-validation results on the English-filtered Twitter partition ($n = 16{,}587$). Values show mean $\pm$ standard deviation across folds.}
\label{tab:classifier_full}
\begin{tabularx}{\textwidth}{@{} l X X X X X @{}} 
\toprule
\textbf{Model} & \textbf{Acc.} & \textbf{Macro F1} & \textbf{Sarc. F1} & \textbf{AUC} & \textbf{MCC} \\
\midrule
GEMA     & $.986{\pm}.006$ & $.985{\pm}.007$ & $.982{\pm}.008$ & $.999{\pm}.000$ & $.970{\pm}.013$ \\
BERTweet & $.975{\pm}.002$ & $.974{\pm}.002$ & $.967{\pm}.002$ & $.997{\pm}.001$ & $.948{\pm}.003$ \\
RoBERTa  & $.991{\pm}.001$ & $.991{\pm}.001$ & $.988{\pm}.001$ & $.999{\pm}.000$ & $.982{\pm}.002$ \\
\bottomrule
\end{tabularx}
\end{table*} 

All four correlations are weakly negative or near zero and lack individual statistical significance. This confirms that $\Delta$ measures a distinct phenomenon: while decontextualised ambiguity captures crowdworker variation in isolation and dictionary polysemy reflects lexicographic sense enumeration, $\Delta$ specifically quantifies contextualised topical coherence in practical tweet usage.

\begin{table}[h]
\centering
\small
\caption{Sanity checks against decontextualised ambiguity \cite{czestochowska2022contextfree} and dictionary polysemy \cite{wijeratne2017emoji_similarity}.}
\label{tab:sanity_full}
\begin{tabularx}{\linewidth}{@{} l X c c c @{}}
\toprule
\textbf{Source} & \textbf{Test} & \textbf{$n$} & \textbf{$\rho$} & \textbf{$p$} \\
\midrule
Czestochowska & $\Delta$ vs.\ description entropy & 59 & $-0.087$ & 0.51 \\
Czestochowska & $\Delta$ vs.\ unique words       & 59 & $-0.029$ & 0.83 \\
\midrule
EmojiNet      & $\Delta$ vs.\ sense count        & 62 & $-0.119$ & 0.36 \\
EmojiNet      & $\Delta$ vs.\ keyword count      & 62 & $+0.075$ & 0.56 \\
\bottomrule
\end{tabularx}
\end{table}

\section{Mean Sentiment Scores}

Table~\ref{tab:placement-sentiment} presents the numerical counterpart to Figure~\ref{fig:violin-placement}.

\begin{table}[h]
\centering
\caption{Mean sentiment scores by emoji position (VADER and TextBlob, scored on
emoji-free text).}
\label{tab:placement-sentiment}
\begin{tabular}{lrrr}
\toprule
\textbf{Position} & \textbf{Count} & \textbf{VADER} & \textbf{TextBlob} \\
\midrule
Start  & $2,188$ & $+0.267$ & $+0.190$ \\
Middle & $4,728$ & $+0.0667$ & $+0.116$ \\
End    & $9,659$ & $-0.018$ & $+0.032$ \\
\bottomrule
\end{tabular}
\end{table}

\section{GEMA Architecture}

The in-depth architectural components and training hyperparameters breakdown for the GEMA model in Section \ref{sec:sarcasm_classifier} is available in Table~\ref{tab:gema_specs}.

\begin{table}[htbp]
  \centering
  \small
  \caption{Architecture components and training hyperparameters for GEMA.}
  \label{tab:gema_specs}
  \begin{tabularx}{\linewidth}{@{} l X @{}}
    \toprule
    \textbf{Component / Parameter} & \textbf{Specification / Value} \\
    \midrule
    \multicolumn{2}{@{}l}{\textbf{Architecture \& Branches}} \\
    Text Encoder & BiGRU over MiniLM embeddings, BERT \texttt{[CLS]} \\
    Emoji Branch & Mean Kralj Novak sentiment, normalized emoji counts \\
    Structural Branch (11-D) & VADER compound (emoji-free), polarity mismatch, caps ratio, punctuation density, norm.\ length, 5 emoji-position indicators, centroid distance\textsuperscript{*} \\
    Branch Fusion & Sigmoid gating \\
    \midrule
    \multicolumn{2}{@{}l}{\textbf{Training Hyperparameters}} \\
    Optimizer & AdamW ($\text{LR} = 10^{-3}$, weight decay $= 10^{-4}$) \\
    Loss Function & Focal Loss ($\alpha = 0.4$, $\gamma = 2.0$) \\
    Batch Size & 16 \\
    Max Epochs / Early Stop & 8 epochs (patience = 2 on validation macro F1) \\
    \bottomrule
    \multicolumn{2}{@{}p{\linewidth}}{\footnotesize \textsuperscript{*}Per-emoji sarcastic vs.\ non-sarcastic distance computed inside each fold.}
  \end{tabularx}
\end{table}

\section{Compute} 
All experiments were run on a CPU. Taking a Ryzen 9 5900HS CPU as the baseline, regenerating every paper number through the code takes about three minutes and the leakage-free five-fold classifier evaluation takes about two hours.

\section{Complete Sarcasm Classifier Performance}
\label{sec:appendix_classifier}

Table~\ref{tab:classifier_full} provides the complete five-metric evaluation across five-fold stratified cross-validation on the English-filtered Twitter partition ($n = 16{,}587$, random state 42).

\section{Data and Code Availability}

The Emoji Sarcasm Dataset~\cite{subramanian2019esd} is released for non-commercial research use and is available at \url{https://github.com/jsubram/Sarcasm-Detection-Using-Emoji}. The Kralj Novak 2015 lexicon~\cite{kraljevnovak2015} comes via the open-source \texttt{emosent-py} package (MIT) \url{https://pypi.org/project/emosent-py/}. Ferr\'{e} 2025 Emoji-Dis~\cite{ferre2025emojidis} is on OSF under CC-BY 4.0 \url{https://osf.io/wv7ub/overview}, Czestochowska 2022 Context-Free Ambiguity~\cite{czestochowska2022contextfree} is on GitHub under MIT \url{https://github.com/epfl-dlab/emoji-ambiguity}, and EmojiNet~\cite{wijeratne2017emoji_similarity} is on Kaggle under CC BY-NC-SA 4.0 \url{https://www.kaggle.com/datasets/rtatman/emojinet}.

Our analysis pipeline, statistical battery, and reproducibility scripts are at the anonymous repository \url{https://anonymous.4open.science/r/StableContextsMixedFeelings/};

\end{document}